\documentclass{article}

\PassOptionsToPackage{numbers}{natbib}

\usepackage[final]{neurips_2018}

\usepackage[utf8]{inputenc} % allow utf-8 input
\usepackage[T1]{fontenc}    % use 8-bit T1 fonts
\usepackage{hyperref}       % hyperlinks
\usepackage{url}            % simple URL typesetting
\usepackage{booktabs}       % professional-quality tables
\usepackage{amsfonts}       % blackboard math symbols
\usepackage{nicefrac}       % compact symbols for 1/2, etc.
\usepackage{microtype}      % microtypography
\usepackage{graphicx}
\usepackage{subcaption}
\usepackage{multirow}
\usepackage{amsmath}
\usepackage{todonotes}

\title{Adaptive Scheduling for Multi-Task Learning}

\author{
  Sébastien~Jean\thanks{Work done while at Google AI.} \\
  Department of Computer Science\\
  New York University \\
  \texttt{sebastien@cs.nyu.edu} \\
  \And
  Orhan Firat \\
  Google AI \\  
  \texttt{orhanf@google.com} \\
  \And
  Melvin Johnson \\
  Google AI \\  
  \texttt{melvinp@google.com} \\
}

\begin{document}
%\nipsfinalcopy is no longer used

\maketitle

\begin{abstract}

To train neural machine translation models simultaneously on multiple tasks (languages), it is common to sample each task uniformly or in proportion to dataset sizes. As these methods offer little control over performance trade-offs, we explore different task scheduling approaches. We first consider existing  non-adaptive techniques, then move on to adaptive schedules that over-sample tasks with poorer results compared to their respective baseline. As explicit schedules can be inefficient, especially if one task is highly over-sampled, we also consider implicit schedules, learning to scale learning rates or gradients of individual tasks instead. These techniques allow training multilingual models that perform better for low-resource language pairs (tasks with small amount of data), while minimizing negative effects on high-resource tasks.
\end{abstract}

\section{Introduction}
 
%emphasize the scale of mt and difficulty.

Multiple tasks may often benefit from others by leveraging more available data. For natural language tasks, a simple approach is to pre-train embeddings~\cite{mikolov2013distributed, peters2018deep} or a language model ~\cite{radford2018improving, devlin2018bert} over a large corpus. The learnt representations may then be used for upstream tasks such as part-of-speech tagging or parsing, for which there is less annotated data. Alternatively, multiple tasks may be trained simultaneously with either a single model or by sharing some model components. In addition to potentially benefit from multiple data sources, this approach also reduces the memory use. However, multi-task models of similar size as single-task baselines often under-perform because of their limited capacity.  The underlying multi-task model learns to improve on harder tasks, but may hit a plateau, while  simpler (or data poor) tasks can be over-trained (over-fitted). Regardless of data complexity, some tasks may be forgotten if the schedule is improper, also known as \textit{catastrophic forgetting}~\cite{FRENCH1999128}.
%Moreover, while the underlying multi-task model learns to improve on harder (or data rich) tasks, simpler (or data poor) tasks appear to be over-trained (over-fitted) or even worse forgotten a.k.a. catastrophic forgetting.

In this paper, we consider multilingual neural machine translation (NMT), where both of the above pathological learning behaviors are observed, sub-optimal accuracy on high-resource, and forgetting on low-resource language pairs.  Multilingual NMT models are generally trained by mixing language pairs in a predetermined fashion, such as sampling from each task uniformly ~\cite{dong2015multi} or in proportion to dataset sizes~\cite{luong2015multi}. While results are generally acceptable with a fixed schedule, it leaves little control over the performance of each task. We instead consider adaptive schedules that modify the importance of each task based on their validation set performance. The task schedule may be modified explicitly by controlling the probability of each task being sampled. Alternatively, the schedule may be fixed, with the impact of each task controlled by scaling the gradients or the learning rates. In this case, we highlight important subtleties that arise with adaptive learning rate optimizers such as Adam~\cite{kingma2014adam}. Our proposed approach improves the low-resource pair accuracy while keeping the high resource accuracy intact within the same multi-task model.
 
\section{Explicit schedules}

A common approach for multi-task learning is to train on each task uniformly~\cite{dong2015multi}. Alternatively, each task may be sampled following a fixed non-uniform schedule, often favoring either a specific task of interest or tasks with larger amounts of data~\cite{luong2015multi, kiperwasser2018scheduled}. Kipperwasser and Ballesteros ~\cite{kiperwasser2018scheduled} also propose variable schedules that increasingly favor some tasks over time. As all these schedules are pre-defined (as a function of the training step or amount of available training data), they offer limited control over the performance of all tasks. 
As such, we consider adaptive schedules that vary based on the validation performance of each task during training.

%\todo{formalization}

To do so, we assume that the baseline validation performance of each task, if trained individually, is known in advance\footnote{Baseline scores can be obtained from already trained single task models, or can be set to an expected value to be reached by the multi-task model.}. When training a multi-task model, validation scores are continually recorded in order to adjust task sampling probabilities. %In the first weighting scheme we consider, the unnormalized weight $w_i$ of task $i$ is given by
%\begin{equation}
%    w_{i} = 1 + \epsilon - \min \left(1, \frac{s_i}{b_i}\right),
%\end{equation}
%where $s_i$ is the latest validation BLEU score and $b_i$ is the baseline performance. Task scores are then normalized into a proper probability distribution.
%As preliminary experiments showed only moderate variation in task probabilities during training, we also explore another variant where unnormalized task weights are set to
The unnormalized score $w_i$ of task $i$ is given by
\begin{equation}
    w_{i} = 1 \bigg/\left(\min \left(1, \frac{s_i}{b_i}\right)^\alpha + \epsilon\right)
\end{equation}
where $s_i$ is the latest validation BLEU score and $b_i$ is the (approximate) baseline performance. Tasks that perform poorly relative to their baseline will be over-sampled, and vice-versa for language pairs with good performance. The hyper-parameter $\alpha$ controls how agressive oversampling is, while $\epsilon$ prevents numerical errors and slightly smooths out the distribution. Final probabilities are simply obtained by dividing the raw scores by their sum.

%\todo{explanation of eq(2)}

\section{Implicit schedules}

Explicit schedules may possibly be too restrictive in some circumstances, such as models trained on a very high number of tasks, or when one task is sampled much more often than others. Instead of explicitly varying task schedules, a similar impact may be achieved through learning rate or gradient manipulation. For example, the GradNorm~\cite{chen2017gradnorm} algorithm scales task gradients based on the magnitude of the gradients as well as on the training losses.

As the training loss is not always a good proxy for validation and test performance, especially compared to a single-task baseline, we continue using validation set performance to guide gradient scaling factors. Here, instead of the previous weighting schemes, we consider one that satisfies the following desiderata. In addition to favoring tasks with low relative validation performance, we specify that task weights are close to uniform early on, when performance is still low on all tasks. We also as set a minimum task weight to avoid catastrophic forgetting.

Task weights $w_i, i=1,...,N$, follow
\begin{equation}
    w_{i} = 1 + (\text{sign } (\overline{S} - S_i))  \min \left(\gamma, \left(\max_j S_j\right)^\alpha \left\vert S_i - \overline{S} \right\vert ^\beta \right),
\label{eq:implicit_val_based}
\end{equation}
\noindent where $S_i = \frac{s_i}{b_i}$ and $\overline{S}$ is the average relative score $(\sum_{j=1}^N S_j)/N$. $\gamma$ sets the floor to prevent catastrophic forgetting, $\alpha$ adjusts how quickly and strongly the schedule may deviate from uniform, while a small $\beta$ emphasizes deviations from the mean score. With two tasks, the task weights already sum up to two, as in GradNorm~\cite{chen2017gradnorm}. With more tasks, the weights may be adjusted so their their sum matches the number of tasks. %Following GradNorm~\cite{chen2017gradnorm}, these weights are adjusted so that their sum is equal to the number of tasks.

\subsection{Optimization details}

Scaling either the gradients $g_t$ or the per-task learning rates $\alpha$ is equivalent with standard stochastic gradient descent, but not with adaptive optimizers such as Adam~\cite{kingma2014adam}, whose update rule is given in Eq.~\ref{eq:adam}.

\begin{equation}
\begin{split}
\hat{m}_t &= (\beta_1 m_{t-1} + (1 - \beta_1) g_t) / (1 - \beta_1^t) \\
\hat{v}_t &= (\beta_2 v_{t-1} + (1 - \beta_2) g_t^2) / (1 - \beta_2^t) \\
\theta_t &= \theta_{t-1} - \alpha \hat{m}_t / (\sqrt{\hat{v}_t } + \epsilon)
\end{split}
\label{eq:adam}
\end{equation}

Moreover, sharing or not the optimizer accumulators (eg. running average of 1\textsuperscript{st} and 2\textsuperscript{nd} moment $\hat{m}_t$ and $\hat{v}_t$ of the gradients) is also impactful.
%\todo{maybe add the formulation} Done
%\todo{refer to adam equation + explain symbols?}
Using separate optimizers and simultaneously scaling the gradients of individual tasks is ineffective. Indeed, Adam is scale-insensitive because the updates are divided by the square root of the second moment estimate $\hat{v}_t$. The opposite scenario, a shared optimizer across tasks with scaled learning rates, is also problematic as the momentum effect ($\hat{m}_t$) will blur all tasks together at every update. All experiments we present use distinct optimizers, with scaled learning rates. The converse, a shared optimizer with scaled gradients, could also potentially be employed.

% Do we have results for implicit schedules with the same task weights as explicit schedules?

%TODO: Talk about Multi-Task Learning Using Uncertainty ... (Kendall et al.) and End-to-End Multi-Task Learning with Attention (Liu et al.). I did not pay attention to latter during internship. They use a weighting scheme based on the loss ratio (L_t / L_{t-1}).

%TODO Catastrophic forgetting citation
%modify model itself.

\section{Experiments}

\subsection{Data}

We extract data from the WMT'14 English-French (En-Fr) and English-German (En-De) datasets. To create a larger discrepancy between the tasks, so that there is a clear dataset size imbalance, the En-De data is artificially restricted to only 1 million parallel sentences, while the full En-Fr dataset, comprising almost 40 million parallel sentences, is used entirely. Words are split into subwords units with a joint vocabulary of 32K tokens.\footnote{Joint vocabulary is extracted from the full En-De and En-Fr datasets.} BLEU scores are computed on the tokenized output with \textit{multi-bleu.perl} from Moses~\cite{koehn2007moses}.

%We extract data from the WMT'14 En-Fr and En-De datasets. To create a larger discrepancy between the tasks, so that there is a clear dataset size imbalance delimitation, the En-De data is artificially restricted to only 1 million parallel sentences, while the full En-Fr, comprising almost 40 million parallel sentences, is used entirely. Words are split into subwords units with a joint vocabulary of 32K tokens.\footnote{Joint vocabulary is extracted from the full En-De and En-Fr datasets.} BLEU scores are computed on the tokenized output with \textit{multi-bleu.perl} from Moses~\cite{koehn2007moses}.

\subsection{Models}

All baselines are Transformer models in their base configuration~\cite{vaswani2017attention}, using 6 encoder and decoder layers, with model and hidden dimensions of 512 and 2048 respectively, and 8 heads for all attention layers. For initial multi-task experiments, all model parameters were shared~\cite{johnson2017google}, but performance was down by multiple BLEU points compared to the baselines. As the source language pair is the same for both tasks, in subsequent experiments, only the encoder is shared~\cite{dong2015multi}. For En-Fr, 10\% dropout is applied as in~\cite{vaswani2017attention}. After observing severe overfitting on En-De in early experiments, the rate is increased to 25\% for this lower-resource task. All models are trained on 16 GPUs, using Adam optimizer with a learning rate schedule (inverse square root~\cite{vaswani2017attention}) and warmup.

%All baselines are Transformer models in their base configuration, using 6 encoder and decoder layers~\cite{vaswani2017attention}, with model and hidden dimensions being 512 and 2048, having 8 heads for all multi-head attention layers. For initial multi-task experiments, all model parameters were shared~\cite{johnson2017google}, but performance was down by multiple BLEU points. As the source language pair is the same for both tasks, in subsequent experiments, only the encoder is shared~\cite{dong2015multi}. For En-Fr, 10\% dropout is applied as in~\cite{vaswani2017attention}. After observing severe overfitting on En-De in early experiments, the rate is increased to 25\% for this lower-resource task.

%\todo{seb: dropout} %Done
% Shared embeddings?

\subsection{Results}

The main results are summarized in Table~\ref{tbl:results}. Considering the amount of training data, we trained single task baselines for 400K and 600K steps for En-De and En-Fr respectively, where multi-task models are trained for 900K steps after training. All reported scores are the average of the last 20 checkpoints. Within each general schedule type, model selection was performed by maximizing the average development BLEU score between the two tasks.

With uniform sampling, results improve by more than 1 BLEU point on En-De, but there is a significant degradation on En-Fr. Sampling En-Fr with a 75\% probability gives similar results on En-De, but the En-Fr performance is now comparable to the baseline. Explicit adaptive scheduling behaves similarly on En-De and somewhat trails the En-Fr baseline.

% Within each general schedule type, model selection was performed by maximizing the average development BLEU score between the two tasks.

% \usepackage{multirow}
\begin{table}[h!]
\centering
\begin{tabular}{l|cc|cc}
\hline
\multirow{2}{*}{Method}  & \multicolumn{2}{c|}{Task 1 (En-De)} & \multicolumn{2}{c}{Task 2 (En-Fr)} \\ \cline{2-5} 
                         & \multicolumn{1}{c}{Dev}   & Test   & \multicolumn{1}{c}{Dev}  & Test   \\ \hline \hline
En-De Baseline           & 23.58                      & 24.90  & -                         & -      \\
En-Fr Baseline           & -                          & -      & \textbf{34.71}                     & 40.80  \\ \hline \hline
Explicit - Constant (50\% En-Fr)      & \textbf{24.80}                      & 26.14  & 34.25                     & 39.98  \\
Explicit - Constant (75\% En-Fr)      & 24.53                      & 26.16  & 34.56                     & \textbf{41.00}  \\
Explicit - Validation based & 24.67                      & 26.35  & 34.55                     & 40.70  \\ \hline \hline
Implicit - GradNorm      & 24.69                      & \textbf{26.42}  & 34.33                     & 40.28  \\
Implicit - Validation based     & 24.32                      & 25.58  & 34.67                     & 40.89  \\ \hline
\end{tabular}
\caption{Comparison of scheduling methods, measured by BLEU scores. Best results in bold.}
\label{tbl:results}
\end{table}

For implicit schedules, GradNorm performs reasonably strongly on En-De, but suffers on En-Fr, although slightly less than with uniform sampling. Implicit validation-based scheduling still improves upon the En-De baseline, but less than the other approaches. On En-Fr, this approach performs about as well as the baseline and the multilingual model with a fixed 75\% En-Fr sampling probability.

Overall, adaptive approaches satisfy our desiderata, satisfactory performance on both tasks, but an hyper-parameter search over constant schedules led to slightly better results. One main appeal of adaptive models is their potential ability to scale much better to a very large number of tasks, where a large hyper-parameter search would prove prohibitively expensive.

Additional results are presented in the appendix.

% \todo{both: in text, discuss different uniform schedules} Done
%\todo{subsection: instabilities (discussion)} Done in discussion.

% \begin{table}[]
% \centering
% \begin{tabular}{l|ll}

% \hline
% Method                   & Task-1 (En-De) & \multicolumn{1}{l}{Task-2 (En-Fr)} \\ \hline \hline
% %En-De Baseline (lr1.5)           &      23.44/24.95          &  - \\ \hline
% En-De Baseline (lr3)             &      23.58/24.90          &  - \\ \hline
% %En-Fr Baseline (lr1.5)           &       -         & 34.64/41.12 \\ \hline
% En-Fr Baseline (lr3)             &       -         & 34.71/40.80 \\ \hline
%  \hline
% Explicit - Uniform       &      24.80/26.14          & 34.25/39.98                                     \\ \hline
% Explicit - Inverse ratio &      24.67/26.35          & 34.55/40.70                                     \\ \hline \hline
% Implicit - GradNorm      &      24.59/26.44          & 34.29/40.23                                    \\ \hline
% Implicit - Val-based     &      24.32/25.58          & 34.67/40.89                                    \\ \hline
% \end{tabular}
% \caption{Final Results. Dev/Test BLEU scores.}
% \end{table}

\section{Discussion and other related work}

% \todo{Talk about bandits, Graves et al.} Done
% \todo{kendall (brief)} Done
% \todo{Teacher-Student Curriculum Learning (different strategy: train on tasks where you are already making the most progress)} DONE
%\todo{hypergradients} Done

To train multi-task vision models, Liu et al.~\cite{liu2018end} propose a similar \textit{dynamic weight average} approach. Task weights are controlled by the ratio between a recent training loss and the loss at a previous time step, so that tasks that progress faster will be downweighted, while straggling ones will be upweighted. This approach contrasts with the curriculum learning framework proposed by Matiisen et al.~\cite{matiisen2017teacher}, where tasks with faster progress are preferred. Loss progress, and well as a few other signals, were also employed by Graves et al.~\cite{graves2017automated}, which formulated curriculum learning as a multi-armed bandit problem. One advantage of using progress as a signal is that the final baseline losses are not needed. \textit{Dynamic weight average} could also be adapted to employ a validation metric as opposed to the training loss. Alternatively, uncertainty may be used to adjust multi-task weights~\cite{kendall2018multi}.

Sener and Volkun~\cite{sener2018multi} discuss multi-task learning as a multi-objective optimization. Their objective tries to achieve Pareto optimality, so that a solution to a multi-task problem cannot improve on one task without hurting another. Their approach is learning-based, and contrarily to ours, doesn't require a somewhat ad-hoc mapping between task performance (or progress) and task weights. However, Pareto optimality of the training losses does not guarantee Pareto optimality of the evaluation metrics. Xu et al. present \textit{AutoLoss}~\cite{xu2018autoloss}, which uses reinforcement learning to train a controller that determines the optimization schedule. In particular, they apply their framework to (single language pair) NMT with auxiliary tasks.

With implicit scheduling approaches, the effective learning rates are still dominated by the underlying predefined learning rate schedule. For single tasks, hypergradient descent~\cite{baydin2017online} adjusts the global learning rate by considering the direction of the gradient and of the previous update. This technique could likely be adapted for multi-task learning, as long as the tasks are sampled randomly.

Tangentially, adaptive approaches may behave poorly if validation performance varies much faster than the rate at which it is computed.
Figure~\ref{fig:bad_boy} (appendix) illustrates a scenario, with an alternative parameter sharing scheme, where BLEU scores and task probabilities oscillate wildly. As one task is favored, the other is catastrophically forgotten. When new validation scores are computed, the sampling weights change drastically, and the first task now begins to be forgotten.

\section{Conclusion}

We have presented adaptive schedules for multilingual machine translation, where task weights are controlled by validation BLEU scores. The schedules may either be explicit, directly changing how task are sampled, or implicit by adjusting the optimization process. Compared to single-task baselines, performance improved on the low-resource En-De task and was comparable on high-resource En-Fr task.

For future work, in order to increase the utility of adaptive schedulers, it would be beneficial to explore their use on a much larger number of simultaneous tasks. In this scenario, they may prove more useful as hyper-parameter search over fixed schedules would become cumbersome.

\clearpage

\bibliography{schedule}
\bibliographystyle{plain}

\iffalse
\section*{References}

\small

[1] Alexander, J.A.\ \& Mozer, M.C.\ (1995) Template-based algorithms
for connectionist rule extraction. In G.\ Tesauro, D.S.\ Touretzky and
T.K.\ Leen (eds.), {\it Advances in Neural Information Processing
  Systems 7}, pp.\ 609--616. Cambridge, MA: MIT Press.

[2] Bower, J.M.\ \& Beeman, D.\ (1995) {\it The Book of GENESIS:
  Exploring Realistic Neural Models with the GEneral NEural SImulation
  System.}  New York: TELOS/Springer--Verlag.

[3] Hasselmo, M.E., Schnell, E.\ \& Barkai, E.\ (1995) Dynamics of
learning and recall at excitatory recurrent synapses and cholinergic
modulation in rat hippocampal region CA3. {\it Journal of
  Neuroscience} {\bf 15}(7):5249-5262.
\fi

\newpage

\appendix
\section{Impact of hyper-parameters}

In this appendix, we present the impact of various hyper-parameters for the different schedule types.

\begin{figure}[h!]
    \centering
    \begin{subfigure}[b]{0.45\textwidth}
        \includegraphics[scale=0.45]{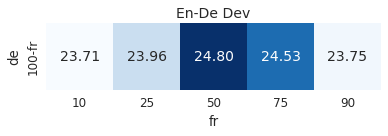}
    \end{subfigure}
    ~
    \begin{subfigure}[b]{0.45\textwidth}
        \includegraphics[scale=0.45]{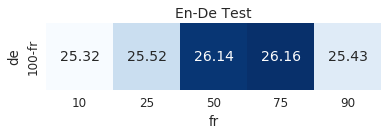}
    \end{subfigure}

    \begin{subfigure}[b]{0.45\textwidth}
        \includegraphics[scale=0.45]{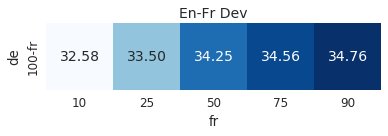}
    \end{subfigure}
    ~
    \begin{subfigure}[b]{0.45\textwidth}
        \includegraphics[scale=0.45]{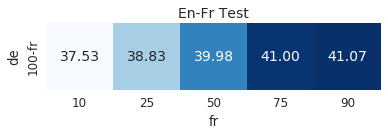}
    \end{subfigure}
    
    \caption{BLEU Score Results for Explicit Constant Schedules. Higher scores are color coded with darker colors and indicate better accuracy.}
    \label{fig:exp_constant}    
\end{figure}

Figure~\ref{fig:exp_constant} illustrates the effect of sampling ratios in explicit constant scheduling. We vary the sampling ratio for a task from 10\% to 90\% and evaluated the development and test BLEU scores by using this fixed schedule throughout the training. Considering the disproportional dataset sizes between two tasks (1/40), oversampling high-resource task yields better overall  performance for both tasks. While a uniform sampling ratio favors the low-resource task (50\%-50\%), more balanced results are obtained with a 75\% - 25\% split favoring the high-resource task. 

Explicit Dev-Based schedule results are illustrated in Figure~\ref{fig:exp_adp_inv} below, where we explored varying $\alpha$ and $\epsilon$ parameters, to control oversampling and forgetting.

\begin{figure}[h!]
    \centering
    \begin{subfigure}[b]{0.45\textwidth}
        \includegraphics[scale=0.45]{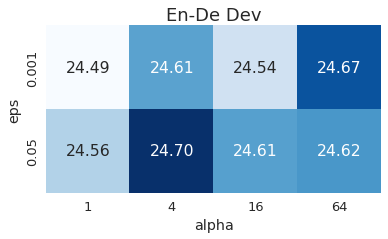}
    \end{subfigure}
    ~ %add desired spacing between images, e. g. ~, \quad, \qquad, \hfill etc. 
    \begin{subfigure}[b]{0.45\textwidth}
        \includegraphics[scale=0.45]{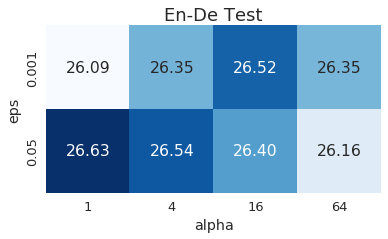}
    \end{subfigure}
    ~ %add desired spacing between images, e. g. ~, \quad, \qquad, \hfill etc. 
    
    \begin{subfigure}[b]{0.45\textwidth}
        \includegraphics[scale=0.45]{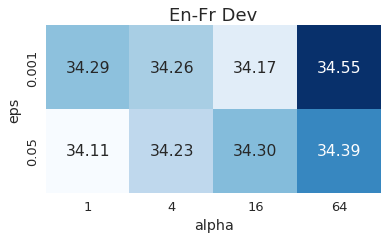}
    \end{subfigure}
    ~ %add desired spacing between images, e. g. ~, \quad, \qquad, \hfill etc. 
    \begin{subfigure}[b]{0.45\textwidth}
        \includegraphics[scale=0.45]{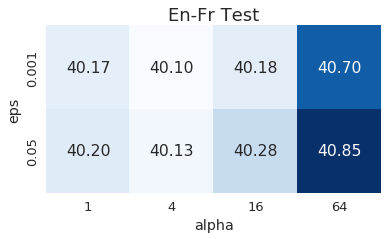}
    \end{subfigure}
    
    \caption{Explicit Dev-Based Schedules}
    \label{fig:exp_adp_inv}    
\end{figure}

%\begin{figure}
%    \begin{subfigure}[b]{0.45\textwidth}
%        \includegraphics[scale=0.33]{pics/bad_explicit_adaptive_inv_dev_sco%res}
%    \end{subfigure} \hfill
%    \begin{subfigure}[b]{0.45\textwidth}
%        \includegraphics[scale=0.33]{pics/bad_explicit_adaptive_inv_task_pr%obs}
%    \end{subfigure}
%    \caption{---}
%    \label{fig:bad_boy}    
%\end{figure}

\begin{figure}[h!]
    \centering
    \begin{subfigure}[b]{0.45\textwidth}
        \includegraphics[scale=0.45]{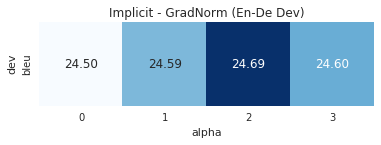}
    \end{subfigure}
    ~ %add desired spacing between images, e. g. ~, \quad, \qquad, \hfill etc. 
    \begin{subfigure}[b]{0.45\textwidth}
        \includegraphics[scale=0.45]{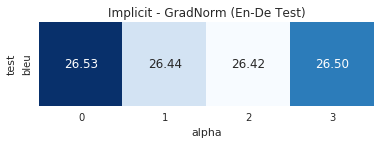}
    \end{subfigure}
    ~ %add desired spacing between images, e. g. ~, \quad, \qquad, \hfill etc. 
    
    \begin{subfigure}[b]{0.45\textwidth}
        \includegraphics[scale=0.45]{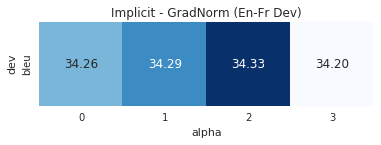}
    \end{subfigure}
    ~ %add desired spacing between images, e. g. ~, \quad, \qquad, \hfill etc. 
    \begin{subfigure}[b]{0.45\textwidth}
        \includegraphics[scale=0.45]{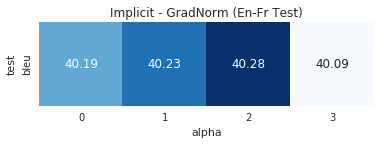}
    \end{subfigure}
    
    \caption{Implicit GradNorm Schedules}
    \label{fig:imp_grad_norm}    
\end{figure}

\begin{figure}[h!]
    \centering
    \begin{subfigure}[b]{0.25\textwidth}
        \includegraphics[scale=0.35]{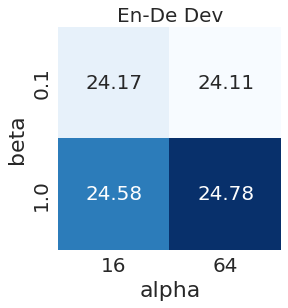}
    \end{subfigure}\hspace*{-0.2em}
    \begin{subfigure}[b]{0.25\textwidth}
        \includegraphics[scale=0.35]{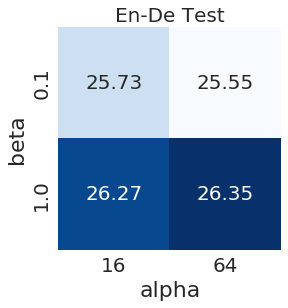}
    \end{subfigure}    \hspace*{-0.5em}
    \begin{subfigure}[b]{0.25\textwidth}
        \includegraphics[scale=0.35]{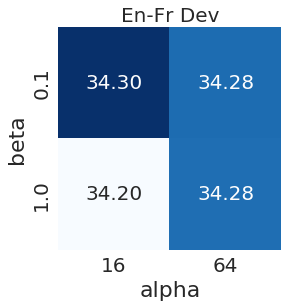}
    \end{subfigure}\hspace*{-0.3em}
    \begin{subfigure}[b]{0.25\textwidth}
        \includegraphics[scale=0.35]{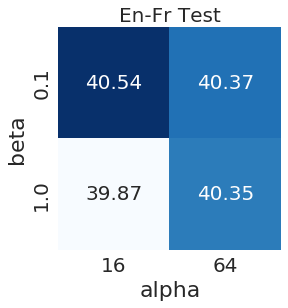}
    \end{subfigure}
    
    \caption{Implicit Dev-Based Schedules}
    \label{fig:imp_dev_based_alphas}    
\end{figure}

\clearpage
\section{Implicit validation-based scheduling progress}

We here present how the task weights, learning rates and validation BLEU scores are modified over time with an implicit schedule. For the implicit schedule hyper-parameters, we set $\alpha=16$, $\beta=0.1$, $\gamma=0.05$ with baselines $b_i$ being 24 and 35 for En-De and En-Fr respectively. For the best performing model, we used inverse-square root learning rate schedule \cite{vaswani2017attention} with a learning rate of 1.5 and 40K warm-up steps.

\begin{figure}[h!]
    \begin{subfigure}[b]{0.45\textwidth}
        \includegraphics[scale=0.33]{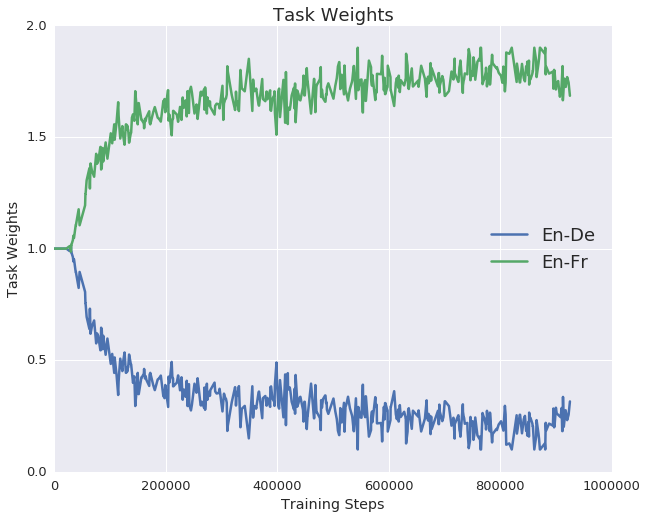}
    \end{subfigure} 
    \begin{subfigure}[b]{0.45\textwidth}
        \includegraphics[scale=0.33]{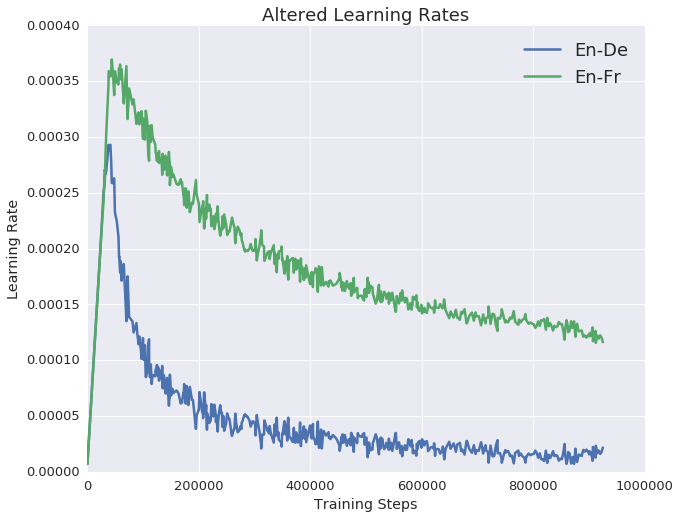}
    \end{subfigure}
    \begin{subfigure}[b]{0.45\textwidth}
        \includegraphics[scale=0.33]{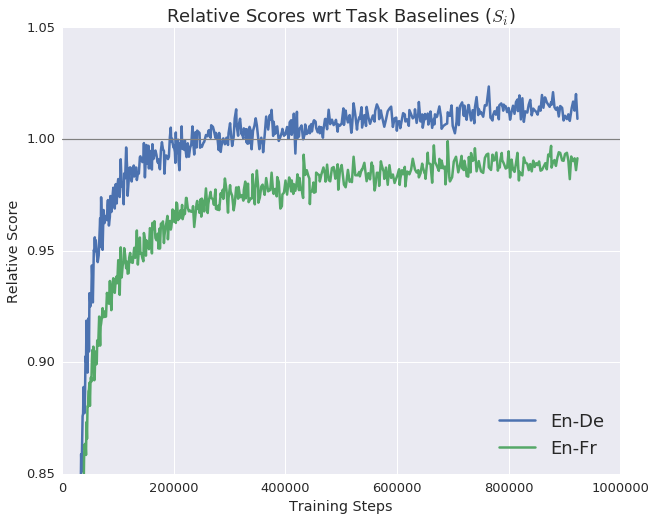}
    \end{subfigure}    
    \hfill
    \begin{subfigure}[b]{0.45\textwidth}
        \includegraphics[scale=0.33]{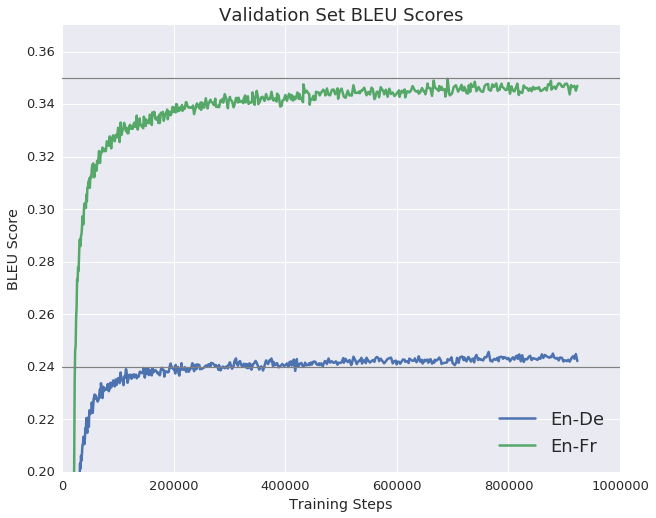}
    \end{subfigure}    
    \caption{Implicit Validation-Based Scheduling Progress. }
    \label{fig:imp_dev_based_curves}    
\end{figure}

Task weights are adaptively changed by the scheduler during training (Figure~\ref{fig:imp_dev_based_curves} top-left), and predicted weights are used to adjust the learning rates for each task (Figure~\ref{fig:imp_dev_based_curves} top-right). Following Eq.~\ref{eq:implicit_val_based}, computed relative scores for each task, $S_j$, are illustrated in Figure~\ref{fig:imp_dev_based_curves} bottom-left. Finally, progression of the validation set BLEU scores with their corresponding baselines (as solid horizontal lines) are given in in Figure~\ref{fig:imp_dev_based_curves} bottom-right.

%\todo{orhan: hyper-params?}

\clearpage
\section{Possible training instabilities}

This appendix presents a failed experiment with wildly varying oscillations. All encoder parameters were tied, as well as the first four layers of the decoder and the softmax. An explicit schedule was employed.

\begin{figure}[h!]
    \begin{subfigure}[b]{0.45\textwidth}
        \includegraphics[scale=0.33]{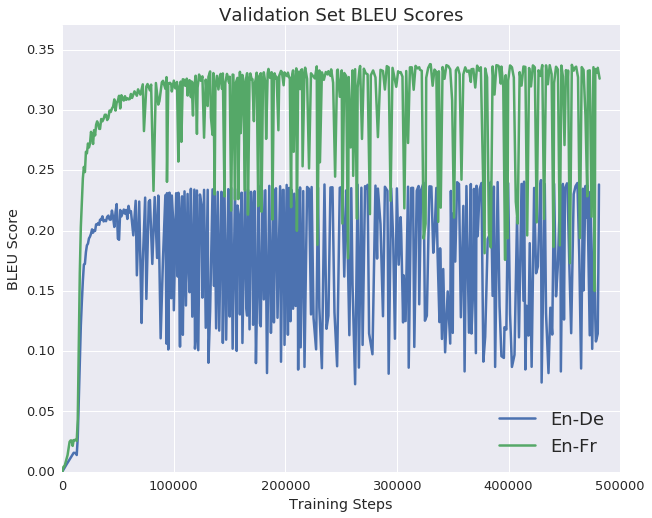}
    \end{subfigure} \hfill
    \begin{subfigure}[b]{0.45\textwidth}
        \includegraphics[scale=0.33]{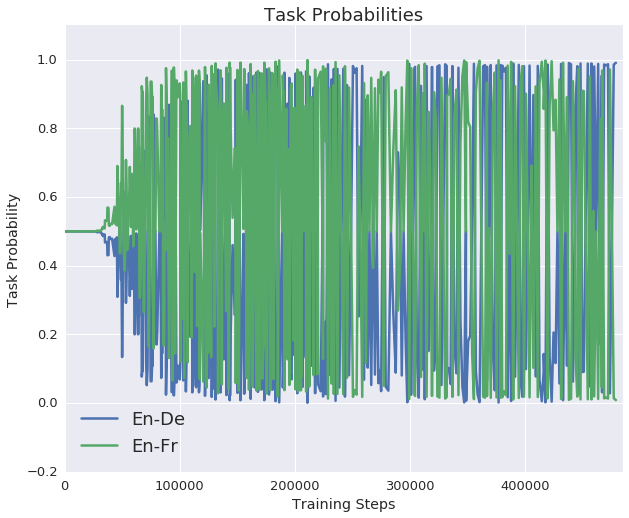}
    \end{subfigure}
    \caption{Wild oscillations}
    \label{fig:bad_boy}    
\end{figure}

\end{document}